\UseRawInputEncoding
\documentclass[runningheads]{llncs}
\usepackage[T1]{fontenc}
\usepackage{graphicx}
\usepackage{amsmath}
\usepackage{subcaption}
\usepackage{hyperref}  

\usepackage{color}

\begin{document}
\title{Pre-training a Transformer-Based Generative Model Using a Small Sepedi Dataset}
\titlerunning{Pre-training GPT model on small dataset}
%
\author{Simon Phetole Ramalepe\inst{1}\orcidID{0000-0001-5660-8099} \and
Thipe I. Modipa\inst{1,3,4}\orcidID{0000-0002-5383-5808} \and
Marelie H. Davel\inst{2,3,4}\orcidID{0000-0003-3103-5858}}

\authorrunning{S. P. Ramalepe et al.}
\institute{Department of Computer Science,
University of Limpopo, South Africa \\
\email{simon.ramalepe@ul.ac.za},  
\email{thipe.modipa@ul.ac.za} 
\and
Faculty of Engineering,
North-West University, South Africa
\and Centre for Artificial Intelligence Research, South Africa
\and National Institute for Theoretical and Computational Sciences, South Africa
}
\maketitle              
\begin{abstract}
Due to the scarcity of data in low-resourced languages, the development of language models for these languages has been very slow. Currently, pre-trained language models have gained popularity in natural language processing, especially, in developing domain-specific models for low-resourced languages. 
In this study, we experiment with the impact of using occlusion-based techniques when training a language model for a text generation task. We curate 2 new datasets, the Sepedi monolingual (SepMono) dataset from several South African resources and the Sepedi radio news (SepNews) dataset from the radio news domain. We use the SepMono dataset to pre-train transformer-based models using the occlusion and non-occlusion pre-training techniques and compare performance. The SepNews dataset is specifically used for fine-tuning. Our results show that the non-occlusion models perform better compared to the occlusion-based models when measuring validation loss and perplexity. However, analysis of the generated text using the BLEU score metric, which measures the quality of the generated text, shows a slightly higher BLEU score for the occlusion-based models compared to the non-occlusion models.

\end{abstract}

\section{Introduction}
Low-resourced languages are characterised by small training datasets, as well as limited tools and linguistic analyses \cite{howcroft_most_2022}. The Sepedi language, one of the official languages in South Africa falls in this category. It is a highly disjunctively written language which is mostly characterised by 2 character words as discussed by the authors in \cite{ramalepe_development_2022}. Previous attempts to develop text-based language models include the use of recurrent neural networks (RNNs) and long-short memory networks (LSTMs)~\cite{sutskever_sequence_2014}. 
These techniques could perform well on short sentences but their performance would degrade as the sentence length increases. The introduction of the transformer architecture~\cite{vaswani_attention_2017} has since gained much attention in developing transformer-based models that can generate longer coherent sentences. This technique uses a multi-head self-attention mechanism to compute the contextual representation of each word by considering the entire input sequence. Complementary approaches when developing language models for low-resourced languages include transfer learning techniques and data augmentation. Transfer learning is a machine learning approach that leverages knowledge gained from pre-trained models to improve the performance of a low-resourced model in a related task \cite{raffel_exploring_2020,wei_overview_2024}. 
In the context of text generation, this approach transfers knowledge from a pre-trained language model to a low-resourced language through the fine-tuning process \cite{howcroft_most_2022,li_pretrained_2021}. Transfer learning has emerged as an approach that effectively addresses data scarcity in low-resourced languages. Data augmentation is another approach that has been explored in several studies~\cite{chang_code-switching_2019,gao_code-switching_2019,pratapa_language_2018,samanta_deep_2019,tarunesh_machine_2021}, with various approaches proposed for generating artificial data. The technique has also been shown to improve the performance of a language model. 
\\
The purpose of this study is to analyse the implications of pre-training a Sepedi generative pre-trained model (GPT) when only a relatively small dataset is available. Specifically, we aim to understand how well a semi-matched dataset (from the same language but a different context) can be used as the seed model for the new context, and whether the type of pre-training used makes a difference. The target context we are interested in is the Sepedi radio news dataset obtained from radio broadcasts, and the semi-matched dataset contains a mix of different styles. 
The study contributes:
\begin{itemize}
    \item Sepedi monolingual (SepMono) dataset curated from several South African resources and Sepedi radio news (SepNews) dataset curated from radio news domain.
    \item A comparison of two different pre-training techniques for a small dataset, before and after fine-tuning.
    \item Sepedi transformer generative pre-trained models (SepGPT) and (SepGPT-OCC), trained on the newly curated Sepedi corpus using the standard GPT training approach and occlusion-based technique.
\end{itemize}


\section{Background}\label{Section:Background-pretraining}

We provide a brief overview of language models and the use of 
pre-training, before discussing selected closely related studies.

\subsection{Pre-trained language models}\label{pre-trained language models}

Language models are applied to many computational linguistic problems including text generation. Language modeling remains the most fundamental task in natural language processing (NLP) and automatic speech recognition systems \cite{winata_multilingual_2021}. It assigns a joint probability distribution over a sequence of linguistic words 
\cite{wei_overview_2024,winata_multilingual_2021}. 
In its simplest form, the joint probability of an entire sequence of words can be estimated by multiplying the number of conditional probabilities using the chain rule.
\begin{equation}
\begin{aligned}
P(w) &= P(w_1, w_2, \dots, w_n) = \prod_{i=1}^n P(w_i \mid w_{<i}) \\
     &= P(w_1) P(w_2 \mid w_1) \dots P(w_i \mid w_1, w_2, \dots, w_{i-1})
\end{aligned}
\label{eq1}
\end{equation}

Pre-trained language models (PLMs) are models that were initially trained on a large corpus to capture broad representations of a language's syntactic and semantic knowledge~\cite{wei_overview_2024}. The primary pre-training objective for generative language models is to predict the missing token in a given sentence. Alternative objectives like ``next-sentence prediction'' are also used for specific purposes, such as understanding sentence relationships. In the ``predicting the missing token'' approach, auto-regressive models like GPT-2 \cite{radford_language_2019} predict the next token based on previous tokens, while Bidirectional encoder representations from transformers (BERT) \cite{devlin_bert_2019}, uses the masked language model (MLM) and next-sentence prediction objectives to predict the masked tokens. The MLM objective is achieved by randomly masking the input tokens and training the model to predict the masked tokens using the rest of the context. The next sentence prediction task is used to train the model to predict the relationship of the next sentence given the previous sentence. We use the terms ''occlusion'' and ''non-occlusion'' to refer to pre-training objectives where some input tokens are occluded or masked and the training process where input tokens are not occluded respectively.  

The occlusion-based pre-training is a common pre-training objective in computer vision \cite{chandel_occlusion_2015}, however, it has been used in generative models~\cite{devlin_bert_2019}. This technique helps the model to be robust to unseen tokens \cite{harbecke_explaining_2021}.  Harbecke et al.~\cite{harbecke_explaining_2021}, show the effectiveness of combining occlusion and language modeling for a classification task. 
In general, occlusion is regarded as an explanation method where the difference in prediction when removing an input feature is seen as an indicator of the importance of the feature \cite{zintgraf_visualizing_2017}. 
On the other hand, the non-occlusion pre-training technique is commonly used in autoregressive models like GPT-2~\cite{radford_language_2019}.

Subsequently, pre-trained models are commonly fine-tuned on smaller datasets to perform several downstream tasks using the learned knowledge. To avoid catastrophic forgetting during the standard fine-tuning approach, Howard et al.~\cite{howard_universal_2018} proposed discriminative fine-tuning (which uses different learning rates for different layers), slanted triangular learning rates (which varies the learning rates using the defined pattern for better convergence), and gradual unfreezing of layers (which unfreezes layers from top to bottom during training). The authors show that these approaches can improve the model's performance.

\subsection{Closely related studies}\label{Section:related work}

Several studies have adopted pre-training before fine-tuning their models on downstream tasks. We discuss some of the most relevant studies in this section, with a specific focus on the size of the dataset and evaluation metrics used to evaluate the performance of the models. 

Niculescu et al.~\cite{niculescu_rogpt2_2021} developed a Romanian text generation model (RoGPT-2) using the GPT-2 architecture. The authors pre-trained their model from scratch using 17GB of monolingual data and fine-tuned it on several tasks for news article generation. 
The performance of their model was evaluated using perplexity as a metric and obtained a score of 34.37 on the validation set and 33.74 on the test set for the natural language generation (NLG) task on their base model. Experimenting with news generation, they obtained a BLEU score of 35.90 between the reference text and the generated text. Similarly,  Buzea et al.~\cite{buzea_automatic_2022} developed a smaller Romanian text generation model (MCBGPT-2) using a standard GPT-2 architecture. The model was trained on a small dataset of 24k news items crawled from online news portals. The performance of the model was monitored using the sparse categorical cross-entropy loss function. 

In another study, Martin et al. \cite{martin_swahbert_2022} developed SwaBERT, a low-resourced language model for the Swahili language using the BERT architecture. The authors trained their model with monolingual data collected from news websites, forums, Wikipedia and popular social media websites with a total size of 105MB. The original BERT's architecture was followed when pre-training their model. In their approach, the authors trained a model using both MLM and next sentence prediction as used in \cite{devlin_bert_2019}. However, during training, they experimented by varying the vocabulary size and the number of training and warm-up steps. The performance of their model was analysed on several downstream tasks including news classification, emotions detection, sentiment analysis and named entity recognition (NER).  

In the study by Wongso et al.~\cite{wongso_pre-trained_2022}, the authors developed several low-resourced models for the Sudanese language. The authors pre-trained GPT-2 Model, BERT and RoBERTa architectures using 785MB monolingual data and evaluated them for emotional classification through fine-tuning. The authors observed that both Sudanese RoBERTa and BERT surpassed or performed comparably to larger multilingual models. Perplexity and validation loss were used to evaluate their models during pre-training. 
The study by Martin et al.~\cite{martin_camembert_2020} also demonstrated that it is possible to pre-train a large language model using a relatively small dataset. The authors pre-trained a GPT-2 model using 4GB of French data. They used the masked language modeling approach as used by RoBERTa as their training objectives. However, instead of fixing the masked token, they introduced dynamic token masking which seemed to improve the model variability and made it robust during training. Their model was evaluated on several downstream tasks including natural language inference. 

Finally, in another low-resource language context, Salim et al.~\cite{salim_banglagpt_2023} developed a BanglaGPT generative model using the GPT architecture. They trained their model from scratch using 26.24GB corpus, scraped from several websites. The model outperformed multilingual GPT (mGPT) and LSTM models with an optimal perplexity score of 2.86. 

\section{Experimental setup}\label{Section: experimental setup-pretraiing}

In this section, we start by describing the datasets used in this study, the model development process, and lastly the evaluation techniques used.

\subsection{Data collection}

\subsubsection{SepMono Dataset}
Table~\ref{monolingual:datasets} lists the available Sepedi monolingual text datasets used to develop the pre-trained models. We indicate the specific size of each dataset to show the relative size and scarcity of the available Sepedi textual datasets.  
We combine all these datasets to create a Sepedi monolingual dataset referred to as ``SepMono''. The SepMono dataset consists of 432,970 sentences with 11,360,000 tokens after cleaning. All of these datasets are freely available for research purposes. 

The  National Centre for Human Language Technology (NCHLT) corpus \cite{puttkammer_nchlt_2014}, was curated from a collection of several South African government entities crawled from gov.za websites. The corpus was collected from various language units from 2007 to 2011. The Autshumato dataset, developed by McKellar~\cite{mckellar_autshumato_2022} was collected from several sources (magazines, policies, newsletters, translation works) and documents crawled from the government domain. 
The News headlines dataset \cite{marivate_investigating_2020}, is another Sepedi text corpus based on radio news headlines. It was crawled from one of the South African national Sepedi radio stations between 2018 and 2020.  The Sepedi newspaper dataset \cite{marivate_vukuzenzele_2023}, was collected from the Vukuzenzele newspaper between 2011 and 2022. The dataset has both monolingual and translated parallel data from English to other South African low-resource languages. The Leipzig dataset \cite{goldhahn_building_2012} consists of newspaper texts and texts randomly collected from the web while Web crawl dataset \cite{conneau_unsupervised_2020,wenzek_ccnet_2020} consists of general data extracted from Common Crawl for various languages including Sepedi. 

\begin{table}[th]
  \centering

\caption{\label{monolingual:datasets} Existing monolingual Sepedi datasets curated to create the SepMono dataset.}
\begin{tabular}{|l|l|} 

\hline
 Dataset (monolingual)&\#Data size\\
 \hline
 NCHLT&12MB\\
 Autshumato1&  19MB\\
 Autshumato2(from bilingual)&16MB \\
 Headlines News &125KB\\
 Vukunzenzele & 600KB\\ 
 Web crawl &8.5MB\\
 Leipzig &9MB\\
 \hline
 
 \hline
 \end{tabular}
\end{table}

We split the dataset into 80\% training, 10\% validation and 10\% testing. We show the size of the partitioned dataset in the first row of Table~\ref{Table:dataset-training}.

\begin{table}[th]
  \centering
  \caption{Datasets partitions used during training and evaluation. The SepMono is used during pre-training process while SepNews-1 and 2 are used for fine-tuning.}
  \label{Table:dataset-training}
\begin{tabular}{|l|l|l|l|l|} 
\hline
\textbf{SepMono Dataset} &\textbf{Training} &\textbf{Validation} &\textbf{Testing}&\textbf{Total data}\\ 
\hline
\#Sentences &346,380& 43,290&43,299&432,970 \\
\#Tokens  &9,020,000& 859,110& 1,480,000& 11,360,000 \\
\#Unique tokens &105,090 & 30,320& 39,720 &125,04 \\
\hline
\textbf{SepNews Dataset}&&&&\\
\hline
\textbf{SepNews-1} & &&&\\
\#Sentences  &12,668 & 3,168  & -&15,836\\ 
\#Tokens  & 357,431&81,896 &-&439,327 \\
\#Unique tokens &14,966&5,612 &-& 16,772\\

\textbf{SepNews-2} &&&&\\
\#Sentences&-&- &3,520&3,520\\
\#Tokens  &-&- &101,758& 101,758\\
\#Unique tokens &-&-&6,420&6,420\\
\hline
\end{tabular}
\end{table}

\subsubsection{SepNews dataset}

The last row of Table \ref{Table:dataset-training} shows the SepNews dataset which represents our target context.
The dataset was curated in two phases (June 2022 to April 2023 and May 2023 to November 2023). We named these datasets ``SepNews-1'' and ``SepNews-2'' respectively. 
The SepNews-1 dataset is split into 80\% training and 20\% validation sets. For testing, we used the SepNews-2 dataset, which comes from a different time period and is therefore a strong proxy for unseen contexts. 

\subsubsection{Data cleaning and pre-processing}\label{data cleaning}
To prepare the data for model training, we remove special characters, repeated full stops, and forward and backward slashes. We also break the sentences at full stops. 
We use the  GPT2TokenizerFast\footnote{\url{https://huggingface.co/transformers/v3.0.2/model\_doc/gpt2.html\#gpt2tokenizerfast}} to tokenized the data. The tokenizer uses the Byte Pair Encoding (BPE) technique which breaks the input text into sub-word tokens based on the most common byte pairs in the dataset. This technique is well suited for morphologically rich languages (such as the Sepedi language) which use a combination of smaller morphological units to construct individual words \cite{mielke_spell_2019}. It also helps to control the vocabulary size during model training. We first normalise the data to lowercase and fine-tune the GPT2TokenizerFast using the SepMono dataset to capture the morphological representation of the Sepedi language.

\subsection{Models}

\subsubsection{Model architecture}
We adopt the GPT-2 architecture~\cite{radford_language_2019} to develop the Sepedi generative model (SepGPT). The GPT-2 architecture uses the decoder part of the transformer architecture~\cite{vaswani_attention_2017} with the pre-training objective of predicting the next word given the previous words. The GPT-2 architecture consists of a series of decoder blocks, each incorporating a masked self-attention block which helps to identify the relevant words the model should focus on. The feed-forward neural network block within the hidden layer, on the other hand, establishes the relationships between the input tokens. At the lower level, it has the token and positional embedding layer to map each token in the vocabulary to a high-dimensional vector representation and to capture the positional embedding of the tokens.  

\subsubsection{Model training}\label{model-training}
The experiments are conducted on Google Colab's cloud development environment\footnote{https://colab.google/notebooks/} with NVIDIA T4 GPUs. We start training the model using the standard pre-training objective of the GPT-2 model and later add the occlusion-based technique to help the model learn different structural representations that capture the semantics of the sentence. To optimise hyperparameters, the Weights and Biases\footnote{https://wandb.ai/site} random sweep configuration is utilized, with validation loss monitored throughout.
The AdamW optimizer is employed, starting with a learning rate of 1e-4, and the model convergence is enhanced using a learning rate scheduler with a warm-up. With a patience of 5, early stopping is applied to control overfitting. Due to limited computational resources, the hyperparameter search is randomized to 20 counts over 100 epochs, exploring the optimal learning rate, number of transformer layers, number of attention heads, and dropout rate (see experimental details in the Appendix for more information). The optimal validation loss and perplexity are recorded at each epoch, while the test loss from the best-performing models is also logged. The non-occlusion model obtained its optimal hyperparameters using a higher number of attention heads and layer blocks, as shown in Table~\ref{table:optimal_hyperparameters} in the Appendix.

\subsubsection{Occlusion-based training}

Occlusion-based techniques have been used extensively in computer vision and classification tasks as discussed in Section \ref{pre-trained language models}. In this study, we experiment with this technique for a text generation task. To train the model using the occlusion-based technique, we use the same experimental setup as described above to find the optimal hyperparameters. However, we now combine the generative approach of predicting the next word given the previous word with an occlusion-based technique to help the model learn different structural representations that capture the semantics of the sentence. In addition to the standard generative approach, the occlusion probability is added as a hyperparameter to the model. We experiment with probabilities of 0.1, 0.3, and 0.5 and record the model's performance per epoch during training. This technique introduces noise to the input text by randomly occluding some tokens in the input sequence based on a given probability. The model is then trained to predict the occluded tokens and recover them using the surrounding non-occluded tokens.

\subsubsection{Fine-tuning models}

To evaluate the performance of the trained models, we fine-tune them for a text generation task using the radio news (SepNews) dataset. 
To fine-tune the models, the same experimental setup as in the pre-training process is applied. The optimal hyperparameters obtained for each model are used, and the models are initialized with the respective pre-trained weights. The models are then fine-tuned for 50 epochs with early stopping. Specifically, we experiment with the gradual unfreezing fine-tuning technique as discussed and experimented in \cite{howard_universal_2018,raffel_exploring_2020}. This technique works by gradually unfreezing model layers (from top to bottom during training). Importantly, the approach helps the model to retain previous knowledge and avoid catastrophic forgetting during fine-tuning. Initially, only the top 2 layers are unfrozen. We set the unfreeze interval to 2 epochs, meaning that at every 2 epochs, additional layers are progressively unfrozen. This interval, basically specifies how often layers are unfrozen during the training process.

\subsection{Evaluation metrics}\label{section: Evaluation metrics}
Although human evaluation is the most standard method for evaluating text generation language models, it is expensive to execute and the results are difficult to reproduce \cite{celikyilmaz_evaluation_2021}. Model evaluation can either be done intrinsically or extrinsically. Intrinsic evaluation methods that have been used for language modeling include the Bilingual Evaluation Understudy (BLEU) score and perplexity \cite{buzea_automatic_2022,gupta_semi-supervised_2020,niculescu_rogpt2_2021}. Perplexity (PPL), remains the most preferred metric to evaluate the performance of a language model~\cite{lee_towards_2021,winata_multilingual_2021}. 
It is the exponentiation of the entropy, which is the average negative log-likelihood of the true word sequence. That is, for a language model that assigns probabilities to sequences of words, the perplexity $PPL$ of a sequence of words $W$ where $W=w_1$,$w_2$,...$w_n$ can be computed as:
\begin{equation}
\text{PPL}(W) = \exp\left( \frac{1}{n} \sum_{i=1}^n -\log P(w_i \mid w_1, \ldots, w_{i-1}) \right)
\label{equation: Perplexity}
\end{equation}
where $P(w_i|w_1,w_2,...w_{i-1})$ is given by a language model and is the probability assigned by the model to the word given the previous words in the sequence.
We also use the categorical cross-entropy loss function to monitor the model's performance during training.

The BLEU score metric is used to evaluate the quality of the text generated by a model. It does this by comparing the generated text to the reference text. Its computation is based on the precision of n-grams between the generated text and the reference text, along with a brevity penalty to handle shorter text. It is calculated as follows:
\begin{equation}
\text{BLEU} = BP \times \exp\left(\sum_{n=1}^{N} w_n \log p_n\right)
\end{equation}
where \(BP\) is the brevity penalty, which adjusts the score based on the length of the generated text compared to the reference length. The brevity penalty is calculated as follows:
\begin{equation}
BP = 
\begin{cases} 
1 & \text{if } c > r \\
\exp\left(1 - \frac{r}{c}\right) & \text{if } c \leq r 
\end{cases}
\end{equation}
where \(c\) is the length of the generated text, and \(r\) is the length of the reference text.

\section{Results}\label{Section:Results-pretraining}

In this section, we discuss the performance of the 4 developed models by analysing their performance using validation loss and perplexity. We start with the analysis of the pre-training results, before demonstrating the effects of fine-tuning. 

\begin{figure}[th]
    \centering
    
    \begin{subfigure}[b]{0.78\textwidth}
        \centering
        \includegraphics[width=\textwidth]{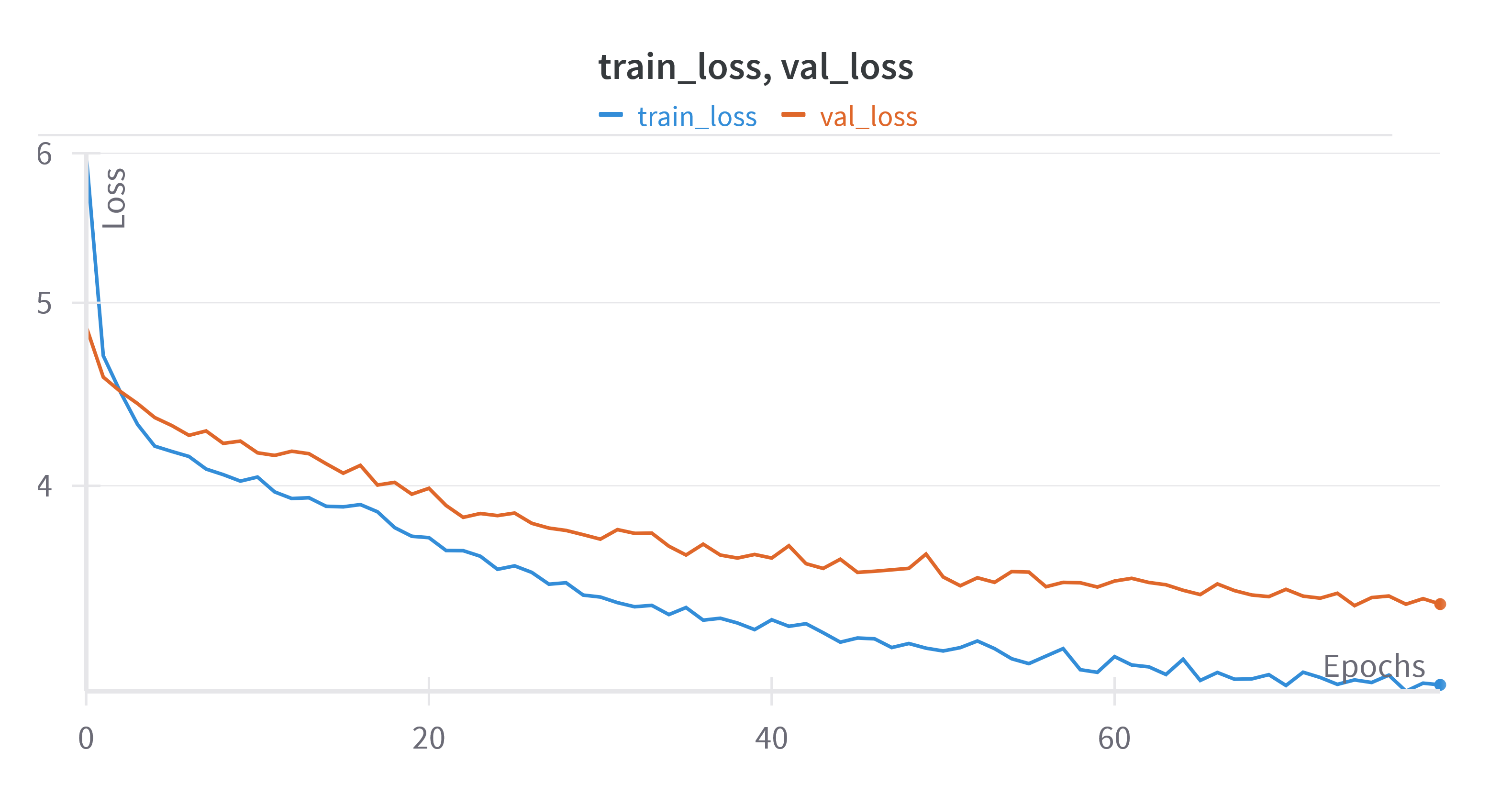}
        \caption{Occlusion-based training curve.}
        \label{fig:occlusion_curve}
    \end{subfigure}
    
    \vspace{1cm}  
    
    \begin{subfigure}[b]{0.78\textwidth}
        \centering
        \includegraphics[width=\textwidth]{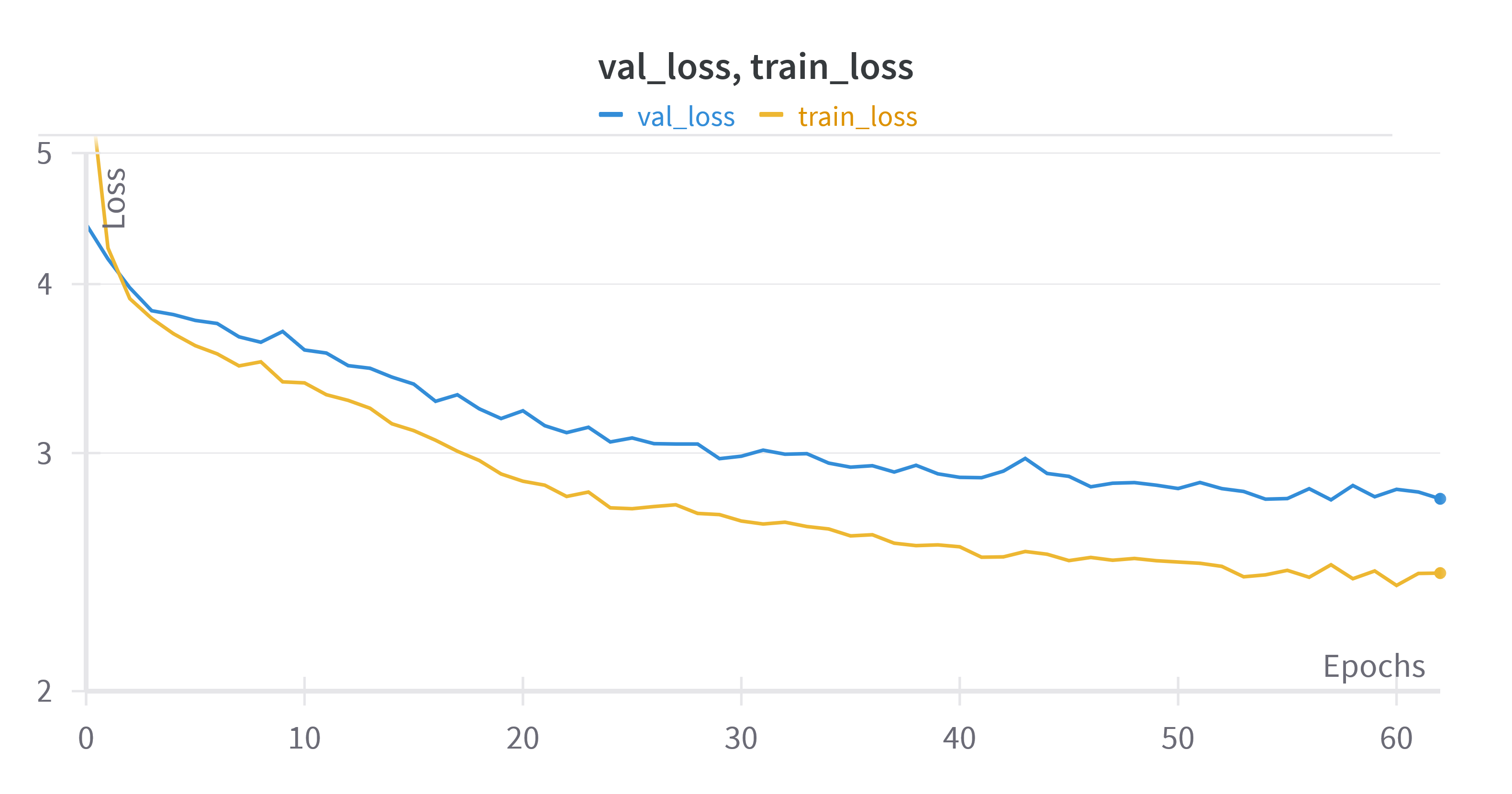}
        \caption{Non-occlusion training curve.}
        \label{fig:non_occlusion_curve}
    \end{subfigure}
    
    \caption{Comparison of training curves: (a) Occlusion-based and (b) Non-occlusion.}
    \label{fig:training_curves}
\end{figure}

Fig.~\ref{fig:occlusion_curve} and \ref{fig:non_occlusion_curve} show the training curves of the optimal scores recorded during the pre-training process of the occlusion and non-occlusion models respectively. Although the maximum number of epochs was set to 100, the occlusion-based model achieved its optimal validation loss of 3.46 at epoch 79, while the non-occlusion model reached its optimal validation loss of 2.78 at epoch 61.
Based on the size of the dataset, and the complexity of occlusion-based training, the SepGPT-OCC obtained the optimal validation loss which is 0.68 higher compared to the non-occlusion model (SepGPT). However, the fine-tuning process (the last row of Table \ref{Table:model-comparison}) significantly closed this gap to just 0.11. Similar observations are noted in validation perplexity, test loss and test perplexity.  Notably, the fine-tuning process improved the validation perplexity of the occlusion-based model by almost 50\%. Of importance to note is that the test loss and test perplexity are relatively high compared to the pre-trained models. The introduction of a completely new test dataset (same news domain but from a different time period) presumably affected the test performance of our fine-tuned models.

We also compare our models with other low-resourced models trained from scratch using the transformer-based technique (the first row of Table \ref{Table:model-comparison}). While systems evaluated on different datasets and for different languages are not directly comparable, this comparison provides an indication of the level of performance achieved. We note that the validation perplexity score of the SepGPT-OCC pre-trained model outperformed both the Sudanese GPT-2 and RoGPT-2 base models by a validation perplexity score of 5.08 and 2.48 respectively. Furthermore, the test perplexity score of the SepGPT-OCC model also outperformed the RoGPT-2 base model. 
Although the optimal test loss obtained from the SepGPT is 2.30 higher compared to the standard BanglaGPT model, the general performance of all our models and the training processes used in this study were satisfactory. 

We further generated text from the trained models and computed a BLEU score to measure the quality of the generated text from the standard SepGPT and the SepGPT-OCC models. Although BLEU score is used mostly in translation tasks, it gives a good indication of similarities between the generated and the reference text. Without human evaluation, the pre-trained SepGPT-OCC model obtained a BLEU score of 5\% higher compared to the SepGPT model while the the BLEU score of the fine-tuned SepGPT-OCC model was 3.86\% higher. These scores show how robust the occlusion model can be to unseen text. We show an example of the generated text from the SepGPT-OCC (FT) model Table \ref{appendix:generated_text} in the Appendix section. Although further analysis of the generated text is necessary we note that the sentences are mostly grammatically correct.

\begin{table}[htbp]
\caption{Model comparisons focusing on validation loss, perplexity, and BLEU score. (FT = Fine-tuned)}
\begin{center}
\resizebox{\textwidth}{!}{  
\begin{tabular}{|l|l|l|l|l|l|l|l|} 
\hline
 
\textbf{Model} & \textbf{Size} & \textbf{Train} & \textbf{Validation} & \textbf{Validation} & \textbf{Test} & \textbf{Test} & \textbf{BLEU} \\ 
 &  & \textbf{Loss} & \textbf{Loss} & \textbf{Perplexity} & \textbf{Loss} & \textbf{Perplexity} & \textbf{Score} \\
\hline
Sudanese GPT-2 & 26GB & 2.43 & 3.61 & 36.97 & - & - & - \\
Sudanese BERT & 26GB & 2.80 & 2.84 & 17.20 & - & - & - \\
Sudanese RoBERTa & 26GB & 1.96 & 1.95 & 7.04 & - & - & - \\
BanglaGPT & 785MB & - & - & 2.86 & 0.45 & - & - \\
RoGPT-2 base & 17GB & - & - & 34.37 & - & 33.74 & 35.90\% \\

\hline
SepGPT-standard & 63MB & 2.44 & 2.78 & 16.04 & 2.75 & 15.71 & 24.19\% \\
SepGPT-occlusion & 63MB & 3.13 & 3.46 & 31.89 & 3.42 & 30.85 & 29.48\% \\
\hline
\hline
SepGPT-standard (FT) & 63MB & 2.28 & 2.69 & 14.87 & 4.02 & 56.00 & 44.98\% \\
SepGPT-occlusion (FT) & 63MB & 2.42 & 2.80 & 16.48 & 4.03 & 56.74 & 48.84\% \\
\hline
\end{tabular}
} 
\label{Table:model-comparison}
\end{center}
\end{table}

Even though the occlusion-based technique introduces noise to the training data, it is observed that both models (pre-trained and fine-tuned) could still produce a competitive result compared to a non-occlusion model. 
Further observations from experiments show that although the performance of the occlusion-based technique is low compared to standard training, the technique can provide valuable insight, especially in building robust code mixed text generation models.

\section{Conclusion}\label{conclusion}\label{Section: Conclusion-pretraining}
The purpose of this study was to evaluate the performance of a generative transformer-based model on a relatively small low-resourced Sepedi dataset. We curated and cleaned 2 datasets, the Sepedi monolingual dataset (SepMono) from various available Sepedi resources and the radio news dataset (SepNews). We used the SepMono dataset to train 2 transformer-based generative models using the standard GPT-2 training objective and the occlusion-based technique. The SepNews dataset was used to fine-tune our 2 pre-trained models using the gradual unfreezing technique. In this process, 2 additional Sepedi text generation models were further developed. Although the transformer-based models are data-intensive, we successfully used the hyperparameter search to obtain the optimal parameters for our pre-trained models (SepGPT and SepGPT-OCC). Compared to other low-resourced GPT models, our models obtained higher and comparable results.
We further used the BLEU score metric to evaluate the performance of the generated text from our trained models and obtain the first and new optimal BLEU score of 44.98\% (SepGPT) and  48.84\% (SepGPT-OCC) for the Sepedi language. The scores obtained in this study create a new baseline for the Sepedi language and other South African low-resourced languages. Although the occlusion-based technique approach obtained in general a higher BLEU Score compared to the non-occlusion model, the non-occlusion model performed fairly better in both validation and perplexity loss which could mean that the model is more reliable in generating coherent text.  
In future work, we aim to experiment with other fine-tuning techniques and also, to analyse the performance of these techniques and their impact on generating code-switched text.

\begin{credits}
\subsubsection{\ackname} We would like to acknowledge the Telkom Centre of Excellence for Speech Technology at the University of Limpopo and the MUST deep learning research group at the Northwest University (Potchefstroom) for their continued support.
This work is based on research supported in part by the National Research Foundation of South Africa (Ref Number RA211019646111).

\subsubsection{\discintname}
The authors have no competing interests to declare that are relevant to the content of this article. 
\end{credits}
\appendix

\renewcommand{\thesection}{}

\section*{Appendix: Experimental Details}
\addcontentsline{toc}{section}{Appendix: Experimental Details}  

\begin{figure}[htbp]
\centering
\includegraphics[width=0.67\textwidth]{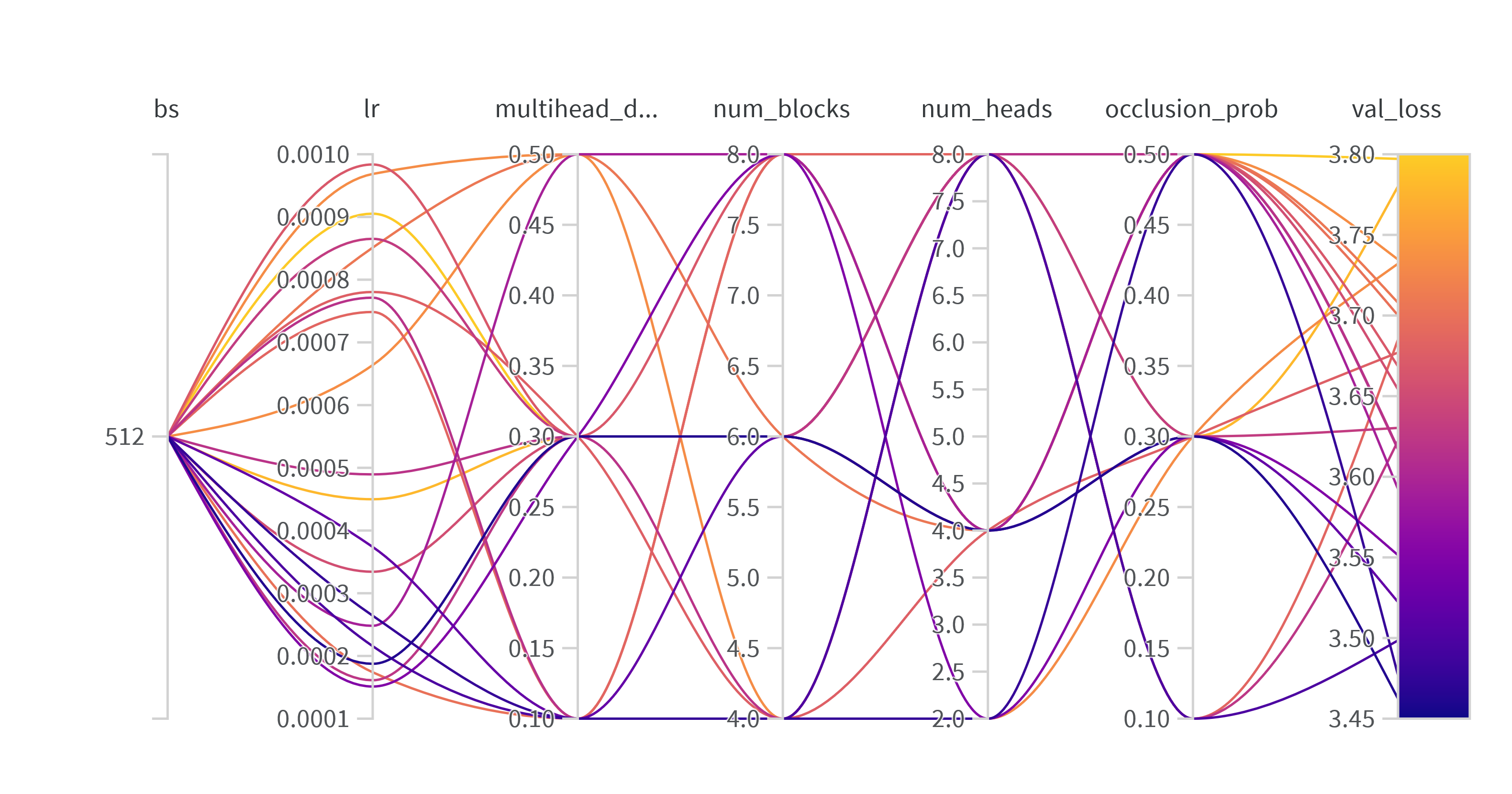}
\caption{Occlusion hyperparameter search}
\label{fig1}
\end{figure}

\label{appendix:detailed_results}  

In Fig.~\ref{fig1}, we show the optimal sweep curves for the hyperparameter search using the occlusion-based approach. For our experiment, the batch size is fixed at 512, and we search for the optimal learning rate, number of transformer layers, number of attention heads, and the dropout rate. The goal is to minimize the validation loss. At each epoch, we record the optimal validation loss and perplexity. Additionally, we include occlusion probabilities as a hyperparameter to evaluate model performance at different occlusion levels. These probabilities are set to 0.1, 0.3, and 0.5.  

We observe from Fig.~\ref{fig1} that when the model becomes too complex, its performance deteriorates (i.e., the lower the validation loss, the better). This could be attributed to the introduction of noise in the training set. Interestingly, the model performed best with an occlusion probability of 0.3.

\vspace{-10pt}  
\begin{table}[htbp]
  \centering
  \caption{Optimal hyperparameters for each model}
  \label{table:optimal_hyperparameters}
  \begin{tabular}{|l|c|c|}
    \hline
    \textbf{Hyperparameter} & \textbf{Occlusion} & \textbf{Non-occlusion} \\
    \hline
    Batch size              & 512 & 512 \\
    Learning rate           & 2e-4 & 2e-4 \\
    Dropout                 & 0.3 & 0.3 \\
    Vocab size              & 50,225 & 50,225 \\
    Weight decay rate       & 1e-2 & 1e-2 \\
    Block layers            & 6 & 8 \\
    Attention heads         & 4 & 8 \\
    Max epochs              & 100 & 100 \\
    Occlusion probability   & 0.3 & - \\
    \hline
  \end{tabular}
\end{table}

\vspace{-10pt}  
\begin{table}[htbp]
\centering
\caption{Translation of Reference and Generated Texts from Sepedi to English}
\label{appendix:generated_text}  
\resizebox{\textwidth}{!}{%
\begin{tabular}{|p{7cm}|p{7cm}|}  
\hline
\textbf{Sepedi Text} & \textbf{English Translation} \\
\hline
Reference Text: moragonyana mothopo woo o akareditše le thuto ya barutwana ya ka sekolong go bala direto tše di ngwadilwego ka maleme a mangwe, ka kudukudu seisimane ga ešita le seafrikaanse. moranang ke ye tee ya dikgwedi tše nne tša goba le matšatši a. morekiši wa dikenywa o ile a... & Reference Text: Later on, that source also included a lesson for students at school to read poems written in other languages, particularly English and Afrikaans. 'Moranang' is one of the four months, having only a few days. A fruit vendor went and... \\
\hline
Generated Text: moragonyana mothopo woo o tsenago kgahlanong le maemong a gagwe ba tsenetše thoko kgopelo ya sekolo kudu gape go šomišana le mašeleng ao a amantšhwago lefaseng ka bophara kudu ke moeta pele wa eff, ka gore mongwaledi phare wa peleng wa naga e lemošitše mošomo: & Generated Text: Later on, that source went against his position, and they joined the request for the school, particularly to collaborate with funds associated worldwide, as led by the former EFF leader, as the former secretary warned of the work. \\
\hline
\end{tabular}
}
\end{table}


%
%
%

\begin{thebibliography}{10}
\providecommand{\url}[1]{\texttt{#1}}
\providecommand{\urlprefix}{URL }
\providecommand{\doi}[1]{https://doi.org/#1}

\bibitem{buzea_automatic_2022}
Buzea, M.C., TrÄƒuÈ™an-Matu, È., Rebedea, T.: Automatic {Romanian} text generation using {GPT}-2. UPB Scientific Bulletin  \textbf{84}(4) (2022)

\bibitem{celikyilmaz_evaluation_2021}
Celikyilmaz, A., Clark, E., Gao, J.: Evaluation of {Text} {Generation}: {A} {Survey} (May 2021)

\bibitem{chandel_occlusion_2015}
Chandel, H., Vatta, S.: Occlusion {Detection} and {Handling}: {A} {Review}. International Journal of Computer Applications  \textbf{120}(10),  33--38 (Jun 2015)

\bibitem{chang_code-switching_2019}
Chang, C.T., Chuang, S.P., Lee, H.Y.: Code-{Switching} {Sentence} {Generation} by {Generative} {Adversarial} {Networks} and its {Application} to {Data} {Augmentation}. In: Interspeech 2019. pp. 554--558. International Speech Communication Association (Sep 2019). \doi{10.21437/Interspeech.2019-3214}

\bibitem{conneau_unsupervised_2020}
Conneau, A., Khandelwal, K., Goyal, N., Chaudhary, V., Wenzek, G., GuzmÃ¡n, F., Grave, E., Ott, M., Zettlemoyer, L., Stoyanov, V.: Unsupervised {Cross}-lingual {Representation} {Learning} at {Scale}. In: Jurafsky, D., Chai, J., Schluter, N., Tetreault, J. (eds.) Proceedings of the 58th {Annual} {Meeting} of the {Association} for {Computational} {Linguistics}. pp. 8440--8451. Online (Jul 2020)

\bibitem{devlin_bert_2019}
Devlin, J., Chang, M.W., Lee, K., Toutanova, K.: {BERT}: {Pre}-training of {Deep} {Bidirectional} {Transformers} for {Language} {Understanding}. In: Proceedings of the 2019 {Conference} of the {North}. pp. 4171--4186. Association for Computational Linguistics, Minneapolis, Minnesota (2019)

\bibitem{gao_code-switching_2019}
Gao, Y., Feng, J., Liu, Y., Hou, L., Pan, X., Ma, Y.: Code-switching sentence generation by {Bert} and {Generative} {Adversarial} {Networks}. In: Interspeech 2019. pp. 3525--3529. International Speech Communication Association (Sep 2019). \doi{10.21437/Interspeech.2019-2501}

\bibitem{goldhahn_building_2012}
Goldhahn, D., Eckart, T., Quasthoff, U.: Building {Large} {Monolingual} {Dictionaries} at the {Leipzig} {Corpora} {Collection}: {From} 100 to 200 {Languages}. In: Calzolari, N., Choukri, K., Declerck, T., DoÄŸan, M.U., Maegaard, B., Mariani, J., Moreno, A., Odijk, J., Piperidis, S. (eds.) Proceedings of the {Eighth} {International} {Conference} on {Language} {Resources} and {Evaluation} ({LREC}'12). pp. 759--765. European Language Resources Association (ELRA), Istanbul, Turkey (May 2012)

\bibitem{gupta_semi-supervised_2020}
Gupta, D., Ekbal, A., Bhattacharyya, P.: A {Semi}-supervised {Approach} to {Generate} the {Code}-{Mixed} {Text} using {Pre}-trained {Encoder} and {Transfer} {Learning}. In: Cohn, T., He, Y., Liu, Y. (eds.) Findings of the {Association} for {Computational} {Linguistics}: {EMNLP} 2020. pp. 2267--2280. Association for Computational Linguistics, Online (Nov 2020). \doi{10.18653/v1/2020.findings-emnlp.206}

\bibitem{harbecke_explaining_2021}
Harbecke, D.: Explaining {Natural} {Language} {Processing} {Classifiers} with {Occlusion} and {Language} {Modeling} (Jan 2021)

\bibitem{howard_universal_2018}
Howard, J., Ruder, S.: Universal {Language} {Model} {Fine}-tuning for {Text} {Classification}. In: Proceedings of the 56th {Annual} {Meeting} of the {Association} for {Computational} {Linguistics} ({Volume} 1: {Long} {Papers}). pp. 328--339. Melbourne, Australia (2018)

\bibitem{howcroft_most_2022}
Howcroft, D.M., Gkatzia, D.: Most {NLG} is {Low}-{Resource}: here's what we can do about it. In: Bosselut, A., Chandu, K., Dhole, K., Gangal, V., Gehrmann, S., Jernite, Y., Novikova, J., Perez-Beltrachini, L. (eds.) Proceedings of the 2nd {Workshop} on {Natural} {Language} {Generation}, {Evaluation}, and {Metrics} ({GEM}). pp. 336--350. Association for Computational Linguistics, Abu Dhabi, United Arab Emirates (Hybrid) (Dec 2022). \doi{10.18653/v1/2022.gem-1.29}

\bibitem{lee_towards_2021}
Lee, N., Bang, Y., Madotto, A., Fung, P.: Towards {Few}-shot {Fact}-{Checking} via {Perplexity}. In: Toutanova, K., Rumshisky, A., Zettlemoyer, L., Hakkani-Tur, D., Beltagy, I., Bethard, S., Cotterell, R., Chakraborty, T., Zhou, Y. (eds.) Proceedings of the 2021 {Conference} of the {North} {American} {Chapter} of the {Association} for {Computational} {Linguistics}: {Human} {Language} {Technologies}. pp. 1971--1981. Association for Computational Linguistics, Online (Jun 2021). \doi{10.18653/v1/2021.naacl-main.158}, \url{https://aclanthology.org/2021.naacl-main.158}

\bibitem{li_pretrained_2021}
Li, J., Tang, T., Zhao, W.X., Wen, J.R.: Pretrained {Language} {Model} for {Text} {Generation}: {A} {Survey}. vol.~5, pp. 4492--4499 (Aug 2021), iSSN: 1045-0823

\bibitem{marivate_vukuzenzele_2023}
Marivate, V., Njini, D., Madodonga, A., Lastrucci, R., Dzingirai, I., Rajab, J.: The {Vuk}'uzenzele {South} {African} multilingual corpus (Feb 2023)

\bibitem{marivate_investigating_2020}
Marivate, V., Sefara, T., Chabalala, V., Makhaya, K., Mokgonyane, T., Mokoena, R., Modupe, A.: Investigating an {Approach} for {Low} {Resource} {Language} {Dataset} {Creation}, {Curation} and {Classification}: {Setswana} and {Sepedi}. In: Mabuya, R., Ramukhadi, P., Setaka, M., Wagner, V., van Zaanen, M. (eds.) Proceedings of the first workshop on {Resources} for {African} {Indigenous} {Languages}. pp. 15--20. European Language Resources Association (ELRA), Marseille, France (May 2020)

\bibitem{martin_swahbert_2022}
Martin, G., Mswahili, M.E., Jeong, Y.S., Woo, J.: {SwahBERT}: {Language} {Model} of {Swahili}. In: Carpuat, M., de~Marneffe, M.C., Meza~Ruiz, I.V. (eds.) Proceedings of the 2022 {Conference} of the {North} {American} {Chapter} of the {Association} for {Computational} {Linguistics}: {Human} {Language} {Technologies}. pp. 303--313. Association for Computational Linguistics, Seattle, United States (Jul 2022). \doi{10.18653/v1/2022.naacl-main.23}, \url{https://aclanthology.org/2022.naacl-main.23}

\bibitem{martin_camembert_2020}
Martin, L., Muller, B., Ortiz~SuÃ¡rez, P.J., Dupont, Y., Romary, L., de~la Clergerie, Ã., Seddah, D., Sagot, B.: {CamemBERT}: a {Tasty} {French} {Language} {Model}. In: Jurafsky, D., Chai, J., Schluter, N., Tetreault, J. (eds.) Proceedings of the 58th {Annual} {Meeting} of the {Association} for {Computational} {Linguistics}. pp. 7203--7219. Association for Computational Linguistics, Online (Jul 2020). \doi{10.18653/v1/2020.acl-main.645}

\bibitem{mckellar_autshumato_2022}
McKellar, C.: Autshumato {Monolingual} {Sepedi} {Corpus} (Sep 2022), accepted: 2022-12-15T06:35:19Z Artwork Medium: Text; UTF8 Interview Medium: Text; UTF8 Publisher: CTexTÂ® (Centre for Text Technology, North-West University)

\bibitem{mielke_spell_2019}
Mielke, S.J., Eisner, J.: Spell {Once}, {Summon} {Anywhere}: {A} {Two}-{Level} {Open}-{Vocabulary} {Language} {Model}. Proceedings of the AAAI Conference on Artificial Intelligence  \textbf{33}(01),  6843--6850 (Jul 2019)

\bibitem{niculescu_rogpt2_2021}
Niculescu, M.A., Ruseti, S., Dascalu, M.: {RoGPT2}: {Romanian} {GPT2} for {Text} {Generation}. In: 2021 {IEEE} 33rd {International} {Conference} on {Tools} with {Artificial} {Intelligence} ({ICTAI}). pp. 1154--1161. IEEE, Washington, DC, USA (Nov 2021). \doi{10.1109/ICTAI52525.2021.00183}

\bibitem{pratapa_language_2018}
Pratapa, A., Bhat, G., Choudhury, M., Sitaram, S., Dandapat, S., Bali, K.: Language {Modeling} for {Code}-{Mixing}: {The} {Role} of {Linguistic} {Theory} based {Synthetic} {Data}. In: Gurevych, I., Miyao, Y. (eds.) Proceedings of the 56th {Annual} {Meeting} of the {Association} for {Computational} {Linguistics} ({Volume} 1: {Long} {Papers}). pp. 1543--1553. Association for Computational Linguistics, Melbourne, Australia (Jul 2018). \doi{10.18653/v1/P18-1143}

\bibitem{puttkammer_nchlt_2014}
Puttkammer, M., Schlemmer, M., Pienaar, W., Bekker, R.: {NCHLT} {Sepedi} text corpora (May 2014)

\bibitem{radford_language_2019}
Radford, A., Wu, J., Child, R., Luan, D., Amodei, D., Sutskever, I.: Language {Models} are {Unsupervised} {Multitask} {Learners}. openAI blog  \textbf{1}(8), ~9 (Feb 2019)

\bibitem{raffel_exploring_2020}
Raffel, C., Shazeer, N., Roberts, A., Lee, K., Narang, S., Matena, M., Zhou, Y., Li, W., Liu, P.J.: Exploring the {Limits} of {Transfer} {Learning} with a {Unified} {Text}-to-{Text} {Transformer}. Journal of Machine Learning Research  \textbf{21}(140),  1--67 (2020)

\bibitem{ramalepe_development_2022}
Ramalepe, S.P., Modipa, T.I., Davel, M.H.: The development of a {Sepedi} text generation model using transformers. In: Proceedings of {South} {Africa} {Telecommunication} {Networks} and {Applications} {Conference} ({SATNAC}). pp. 51--56. Fancourt, Western Cape, South Africa, (2022)

\bibitem{salim_banglagpt_2023}
Salim, M., Murad, H., Das, D., Ahmed, F.: {BanglaGPT}: {A} {Generative} {Pretrained} {Transformer}-{Based} {Model} for {Bangla} {Language} (Sep 2023). \doi{10.1109/ICICT4SD59951.2023.10303383}

\bibitem{samanta_deep_2019}
Samanta, B., Reddy, S., Jagirdar, H., Ganguly, N., Chakrabarti, S.: A deep generative model for code switched {Text}. Proceedings of the Twenty-Eighth International Joint Conference on Artificial Intelligence pp. 5175--5181 (Aug 2019)

\bibitem{sutskever_sequence_2014}
Sutskever, I., Vinyals, O., Le, Q.V.: Sequence to sequence learning with neural networks. In: Proceedings of the 27th {International} {Conference} on {Neural} {Information} {Processing} {Systems} - {Volume} 2. pp. 3104--3112. {NIPS}'14, MIT Press, Cambridge, MA, USA (Dec 2014)

\bibitem{tarunesh_machine_2021}
Tarunesh, I., Kumar, S., Jyothi, P.: From {Machine} {Translation} to {Code}-{Switching}: {Generating} {High}-{Quality} {Code}-{Switched} {Text} (Jul 2021), arXiv:2107.06483 [cs]

\bibitem{vaswani_attention_2017}
Vaswani, A., Shazeer, N., Parmar, N., Uszkoreit, J., Jones, L., Gomez, A.N., Kaiser, Å., Polosukhin, I.: Attention is all you need. In: Advances in {Neural} {Information} {Processing} {Systems}. vol.~30. Curran Associates, Inc. (2017)

\bibitem{wei_overview_2024}
Wei, C., Wang, Y.C., Wang, B., Kuo, C.C.J.: An {Overview} of {Language} {Models}: {Recent} {Developments} and {Outlook}. APSIPA Transactions on Signal and Information Processing  \textbf{13}(2) (2024). \doi{10.1561/116.00000010}

\bibitem{wenzek_ccnet_2020}
Wenzek, G., Lachaux, M.A., Conneau, A., Chaudhary, V., GuzmÃ¡n, F., Joulin, A., Grave, E.: {CCNet}: {Extracting} {High} {Quality} {Monolingual} {Datasets} from {Web} {Crawl} {Data}. In: Calzolari, N., BÃ©chet, F., Blache, P., Choukri, K., Cieri, C., Declerck, T., Goggi, S., Isahara, H., Maegaard, B., Mariani, J., Mazo, H., Moreno, A., Odijk, J., Piperidis, S. (eds.) Proceedings of the {Twelfth} {Language} {Resources} and {Evaluation} {Conference}. pp. 4003--4012. European Language Resources Association, Marseille, France (May 2020), \url{https://aclanthology.org/2020.lrec-1.494}

\bibitem{winata_multilingual_2021}
Winata, G.I.: Multilingual {Transfer} {Learning} for {Code}-{Switched} {Language} and {Speech} {Neural} {Modeling} (Apr 2021), arXiv:2104.06268 [cs, eess]

\bibitem{wongso_pre-trained_2022}
Wongso, W., Lucky, H., Suhartono, D.: Pre-trained transformer-based language models for {Sundanese}. Journal of Big Data  \textbf{9}(1), ~39 (Apr 2022). \doi{10.1186/s40537-022-00590-7}

\bibitem{zintgraf_visualizing_2017}
Zintgraf, L.M., Cohen, T.S., Adel, T., Welling, M.: Visualizing {Deep} {Neural} {Network} {Decisions}: {Prediction} {Difference} {Analysis}. In: 5th {International} {Conference} on {Learning} {Representations}, {ICLR} 2017, {Toulon}, {France}, {April} 24-26, 2017, {Conference} {Track} {Proceedings}. OpenReview.net (2017)

\end{thebibliography}

\end{document}